\pdfoutput=1
\documentclass[11pt]{article}
\usepackage{utils/utils}
\usepackage{hyperref}   

\usepackage[final]{acl}

\newcommand{\eg}{{\emph{e.g., }}}

\author{
  Wenhao Liu\textsuperscript{1}\thanks{Core contribution.},
  Zhenyi Lu\textsuperscript{2}\footnotemark[1],
  Xinyu Hu\textsuperscript{3}\footnotemark[1],
  Jierui Zhang\textsuperscript{1}\footnotemark[1],
  Dailin Li\textsuperscript{4}\footnotemark[1], \\
  \textbf{Jiacheng Cen\textsuperscript{5}}\footnotemark[1], 
  \textbf{Huilin Cao}\textsuperscript{6}, 
  \textbf{Haiteng Wang}\textsuperscript{7}, 
  \textbf{Yuhan Li}\textsuperscript{8}, 
  \textbf{Kun Xie}\textsuperscript{1},
  \textbf{Dandan Li}\textsuperscript{1}, \\
  \textbf{Pei Zhang}\textsuperscript{1},
  \textbf{Chengbo Zhang}\textsuperscript{9},
  \textbf{Yuxiang Ren}\textsuperscript{10}\thanks{Corresponding authors: Xiaohong Huang and Yuxiang Ren.},
  \textbf{Xiaohong Huang}\textsuperscript{1}\footnotemark[2],
  \textbf{Yan Ma}\textsuperscript{1}\\[6pt]
  \textsuperscript{1}State Key Laboratory of Networking and Switching Technology, \\
  Beijing University of Posts and Telecommunications\\
  \textsuperscript{2}Huazhong University of Science and Technology,
  \textsuperscript{3}University of Science and Technology of China,\\
  \textsuperscript{4}Dalian University of Technology,
  \textsuperscript{5}Renmin University of China,
  \textsuperscript{6}Shanghai Jiao Tong University,\\
  \textsuperscript{7}Beihang University,
  \textsuperscript{8}The Hong Kong University of Science and Technology (Guangzhou),\\
  \textsuperscript{9}Peking University,
  \textsuperscript{10}School of Intelligence Science and Technology, Nanjing University\\
  \texttt{\{wenhaoliu,pat,dandl,zhangpei,huangxh,mayan\}@bupt.edu.cn}
}

\title{STORM-BORN: A Challenging Mathematical Derivations Dataset \\Curated via a Human-in-the-Loop Multi-Agent Framework}

\begin{document}
\maketitle

\begin{abstract}

High-quality math datasets are crucial for advancing the reasoning abilities of large language models (LLMs). However, existing datasets often suffer from three key issues: outdated and insufficient challenging content, neglecting human-like reasoning, and limited reliability due to single-LLM generation.
To address these, we introduce \textbf{STORM-BORN}, an ultra-challenging dataset of mathematical derivations sourced from cutting-edge academic papers, which includes dense human-like approximations and heuristic cues.
To ensure the reliability and quality, we propose a novel human-in-the-loop, multi-agent data generation framework, integrating reasoning-dense filters, multi-agent collaboration, and human mathematicians' evaluations. 
We curated a set of 2,000 synthetic samples and deliberately selected the 100 most difficult problems.
Even most advanced models like GPT-o1 solved fewer than 5\% of them. Fine-tuning on STORM-BORN boosts accuracy by 7.84\% (LLaMA3-8B) and 9.12\% (Qwen2.5-7B).
As AI approaches mathematician-level reasoning, STORM-BORN provides both a high-difficulty benchmark and a human-like reasoning training resource.  Our code and dataset are publicly available at \url{https://github.com/lwhere/STORM-BORN}.

\end{abstract}

\begin{figure*}[t]
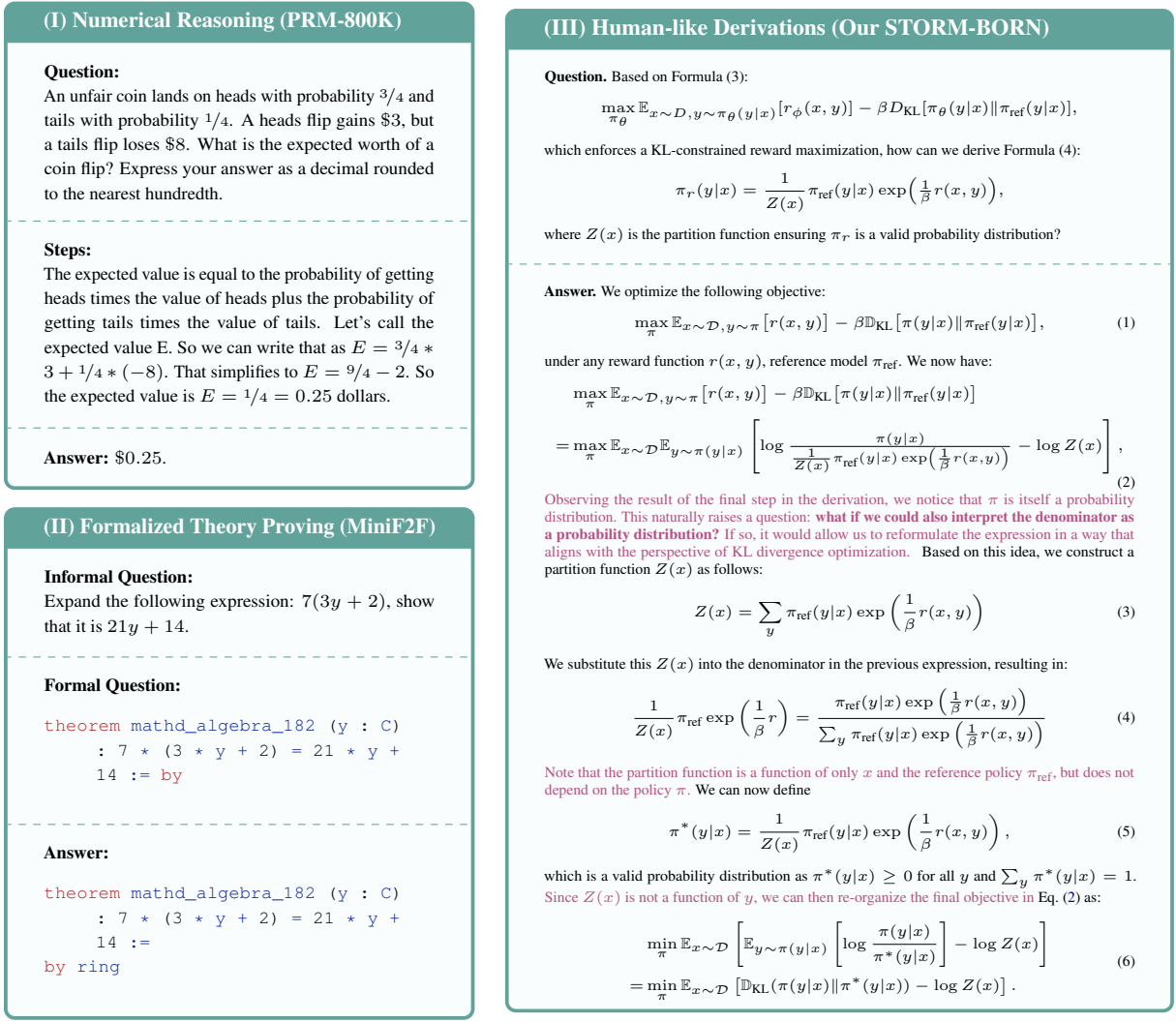

\centering
\begin{minipage}{0.40\textwidth}
\begin{tcolorbox}[
    title=\textbf{\fontsize{8.5}{9.5}\selectfont (I) Numerical Reasoning (PRM-800K)},
    colback=TiffanyBlue!5!white, colframe=TiffanyBlue!75!black,
]
\fontsize{7}{9.5}\selectfont
\textbf{Question: } 

{An unfair coin lands on heads with probability $\nicefrac34$ and tails with probability $\nicefrac14$. A heads flip gains $\$3$, but a tails flip loses $\$8$. What is the expected worth of a coin flip? Express your answer as a decimal rounded to the nearest hundredth.}

\tcbline
\textbf{\fontsize{7}{9.5}\selectfont Steps: } 

{ The expected value is equal to the probability of getting heads times the value of heads plus the probability of getting tails times the value of tails. Let's call the expected value E. So we can write that as $E=\nicefrac{3}{4}*3+\nicefrac{1}{4}*(-8)$. That simplifies to $E=\nicefrac{9}{4}-2$. So the expected value is $E=\nicefrac{1}{4}=0.25$ dollars. }

\tcbline
\textbf{Answer:} $\$0.25$. 
\end{tcolorbox}
\begin{tcolorbox}[
    title=\textbf{\fontsize{8.5}{9.5}\selectfont (II) Formalized Theory Proving (MiniF2F)},
    colback=TiffanyBlue!5!white, colframe=TiffanyBlue!75!black,
]
\fontsize{7}{9.5}\selectfont
\textbf{Informal Question:}\\
Expand the following expression: $7(3y+2)$, show that it is $21y+14$.
\tcbline
\textbf{Formal Question:}
\begin{lstlisting}[basicstyle=\ttfamily]
theorem mathd_algebra_182 (y : C) : 7 * (3 * y + 2) = 21 * y + 14 := by
\end{lstlisting}
\tcbline
\textbf{Answer:} 
\begin{lstlisting}[basicstyle=\ttfamily]
theorem mathd_algebra_182 (y : C) : 7 * (3 * y + 2) = 21 * y + 14 :=
by ring
\end{lstlisting}
\end{tcolorbox}
\end{minipage}
\hfill%
\begin{minipage}{0.58\textwidth}
\resizebox{\linewidth}{!}{
\begin{tcolorbox}[
    title=\textbf{\footnotesize (III) Human-like Derivations (Our STORM-BORN)},
    colframe=lightgray,
    colback=TiffanyBlue!5!white, colframe=TiffanyBlue!75!black,
]
\tiny
\textbf{Question.} Based on Formula (3): 
\vspace{-0.2cm}
\begin{equation*}
    \max _{\pi_{\theta}} \mathbb{E}_{x \sim D, y \sim \pi_{\theta}(y|x)}[r_{\phi}(x, y)]-\beta D_{\text{KL}}[\pi_{\theta}(y | x) \| \piref (y | x)], \vspace{-0.2cm}
\end{equation*}
which enforces a KL-constrained reward maximization, how can we derive Formula (4): 
\vspace{-0.2cm}
\begin{equation*}
    \pi_{r}(y | x)=\frac{1}{Z (x)} \piref (y | x) \exp\Bigl (\tfrac{1}{\beta} r (x, y)\Bigr), \vspace{-0.2cm}
\end{equation*}
where $Z (x)$ is the partition function ensuring $\pi_{r}$ is a valid probability distribution?
\tcbline
\textbf{Answer.} We optimize the following objective:\vspace{-0.2cm}
\begin{equation}
\max_{\pi}  \mathbb{E}_{x\sim \mathcal{D}, y\sim \pi}\bigl[r (x, y)\bigr] - \beta\mathbb{D}_{\textrm{KL}}\bigl[\pi (y|x)\|\piref (y|x)\bigr],\vspace{-0.2cm}
\end{equation}
under any reward function $r (x,y)$, reference model $\piref$. We now have:\vspace{-0.2cm}
\begin{equation}\label{eq:RL_proof}
\begin{aligned}
&\max_{\pi}\textstyle\mathbb{E}_{x\sim \mathcal{D}, y\sim \pi}\bigl[r (x, y)\bigr] - \beta\mathbb{D}_{\textrm{KL}}\bigl[\pi (y|x)\|\piref (y|x)\bigr] \\
=&\max_{\pi}\textstyle\mathbb{E}_{x\sim \mathcal{D}}\mathbb{E}_{y\sim \pi (y|x)}\left[\log\frac{\pi (y|x)}{\frac{1}{Z (x)}\piref (y|x)\exp\left (\frac{1}{\beta}r (x, y)\right)} - \log Z (x)\right],
\end{aligned}
\end{equation}
\textcolor{magenta!70!black}{Observing the result of the final step in the derivation, we notice that $\pi$ is itself a probability distribution. This naturally raises a question: \textbf{what if we could also interpret the denominator as a probability distribution?} If so, it would allow us to reformulate the expression in a way that aligns with the perspective of KL divergence optimization.
}
Based on this idea, we construct a partition function $Z(x)$ as follows:

\begin{equation}
Z(x) = \sum_y \pi_{\text{ref}}(y|x) \exp\left(\frac{1}{\beta} r(x, y)\right)
\end{equation}

We substitute this $Z(x)$ into the denominator in the previous expression, resulting in:

\begin{equation}
\frac{1}{Z(x)} \pi_{\text{ref}} \exp\left(\frac{1}{\beta} r\right) 
= \frac{\pi_{\text{ref}}(y|x) \exp\left(\frac{1}{\beta} r(x, y)\right)}{\sum_y \pi_{\text{ref}}(y|x) \exp\left(\frac{1}{\beta} r(x, y)\right)}
\end{equation}


\textcolor{magenta!70!black}{Note that the partition function is a function of only $x$ and the reference policy $\piref$, but does not depend on the policy $\pi$.} We can now define\vspace{-0.2cm}
\begin{equation}
    \pi^*(y|x) = \frac{1}{Z (x)}\piref (y|x)\exp\left (\frac{1}{\beta}r (x, y)\right),\vspace{-0.2cm}
\end{equation}
which is a valid probability distribution as $\pi^*(y|x)\geq 0$ for all $y$ and $\sum_{y}\pi^*(y|x)=1$. \textcolor{magenta!70!black}{Since $Z (x)$ is not a function of $y$, we can then re-organize the final objective in} \cref{eq:RL_proof} as: \vspace{-0.2cm}
\begin{equation}
\begin{aligned}
&\min_{\pi}  \mathbb{E}_{x\sim \mathcal{D}}\left[\mathbb{E}_{y\sim \pi (y|x)}\left[\log\frac{\pi (y|x)}{\pi^*(y|x)}\right] - \log Z (x)\right]\\
=&\min_{\pi}\mathbb{E}_{x\sim\mathcal{D}}\left[\mathbb{D}_{\text{KL}}(\pi (y|x)\|\pi^*(y|x)) - \log Z (x)\right].
\end{aligned}\vspace{-0.2cm}
\end{equation}

\end{tcolorbox}
}
\end{minipage}%
\caption{
(I) Numerical Reasoning datasets (\eg PRM-800K) require numerical values, which may be too simplistic for advanced LLMs. 
(II) Formalized Theory Proving datasets (\eg MiniF2F) depend on formal languages like Lean, limiting intuitive reasoning and informal generalization.  
(III) In contrast, our STORM-BORN dataset emphasizes human-like reasoning (highlighted in \textcolor{magenta!70!black}{purple}), requiring deep understanding and creativity, more challenging than (I) and more interpretable/generalizable than (II).
}
\label{fig:splash}
\end{figure*}

\section{Introduction}

Mathematical reasoning has emerged as a cornerstone for scaling large language models (LLMs) and probing their upper bounds of intelligence \cite{deepseekmathpushinglimitsmathematical,ye2024physics,glazer2024frontiermathbenchmarkevaluatingadvanced}. Recent advances stem from architectural innovations \cite{mcleish2024transformersarithmeticrightembeddings,li2025large}, enhanced pretraining data \cite{deepseekmathpushinglimitsmathematical,finemath,mathpile}, supervised fine-tuning \cite{yu2024metamathbootstrapmathematicalquestions,cobbe2021trainingverifierssolvemath,fan2025makeloragreatagain}, reinforcement learning \cite{wang2024mathshepherdverifyreinforcellms,zelikman2022starbootstrappingreasoningreasoning}, and chain-of-thought prompting \cite{auto-cot,lu2024mitigatingboundaryambiguityinherent,liu2024optimizing}.
Current supervised mathematical datasets fall into two categories: numerical reasoning focuses on arithmetic computations yielding numbers \cite{cobbe2021trainingverifierssolvemath,math,glazer2024frontiermathbenchmarkevaluatingadvanced}, 
and theorem proving uses formal languages for computer-verifiable proofs \cite{ying2024leanworkbooklargescalelean,wu2024leangithubcompilinggithublean}.

However, existing mathematical datasets face several challenges. \textit{First, they lack nuance and complexity}. 
Current datasets, like GSM8k \cite{cobbe2021trainingverifierssolvemath} and MATH \cite{math}, focus on grade-school or competition problems, offering limited complexity. 
As LLMs excel on these benchmarks, more advanced mathematical reasoning tasks are needed.
\textit{Second, human-like reasoning is limited}. While formal languages \cite{lean} ensure precise verification in recent datasets \cite{ying2024leanworkbooklargescalelean,wu2024leangithubcompilinggithublean}, they obscure interpretable and intuitive reasoning processes \cite{chervonyi2025goldmedalistperformancesolvingolympiad}. 
\textit{Finally, synthetic data lacks reliable annotations}. While LLMs are used to generate large-scale math data \cite{yu2024metamathbootstrapmathematicalquestions,deepseekmathpushinglimitsmathematical,chen2024graphwiz}, they struggle with curating and evaluating expert-level derivations, leading to unreliable step-by-step annotations.

To address these limitations, we focus on a previously underexplored area: \textit{mathematical derivation}. These tasks involve long chains of reasoning (CoT), revisions, and iterative computations, posing significant challenges for current LLMs. Moreover, the data source is both accessible and scalable (\eg academic papers).
Given the complexity of mathematical reasoning, expert annotation is costly and error-prone, while single LLMs often lack reliability. To overcome this, we developed \textit{STORM} (\underline{S}ynergistic \underline{T}heorem and f\underline{OR}mula \underline{M}ining), a multi-agent framework that extracts deduction logic, generates question-answer pairs, and performs iterative refinement to ensure reliability. 

Using STORM, we synthesized 2,000 high-quality math derivation samples.
Recent studies \cite{ye2025limoreasoning,muennighoff2025s1simpletesttimescaling} show that dataset quality and complexity are key drivers of reasoning improvements, especially for reasoning-focused models. 
This is reflected in datasets like Frontier Math (50 samples) \cite{glazer2024frontiermathbenchmarkevaluatingadvanced} and LIMO (817 samples) \cite{ye2025limoreasoning}, where small and rigorously curated datasets lead to significant reasoning gains.
Based on this insight, we further integrate human experts into the curation process, selecting the 100 most challenging problems to form our dataset: \textbf{STORM-BORN}, prioritizing reasoning density and correctness. Notably, even advanced models like Deepseek-R1 and GPT-o1-Pro achieve less than 5\% accuracy on STORM-BORN, compared to 95\% on GSM8K, underscoring its exceptional difficulty.
Furthermore, fine-tuning on STORM-BORN leads to strong generalization on numerical reasoning tasks: TinyLlama achieves a 233\% relative improvement on MATH, and Qwen improves by 16.7\%, despite STORM-BORN not containing explicit numerical reasoning tasks. This demonstrates the broad reasoning skills acquired through our dataset.

Our key contributions can be summarized as follows: 
(1) We introduce a challenging mathematical derivation dataset curated from recent high-impact papers, featuring complex derivations that demand deep theoretical understanding and creative reasoning.
(2) We develop a human-in-the-loop, multi-agent data generation system that extracts complete derivation processes, ensuring human-like reasoning patterns and high-quality, reliable annotations.
(3) Extensive human and automatic assessments confirm that even the most advanced LLMs solve fewer than 5\% of STORM-BORN problems, highlighting its difficulty. Meanwhile, fine-tuning on our dataset yields strong generalization, particularly for numerical reasoning tasks.

\section{Related Work}

\subsection{Large Language Models for Mathematical Reasoning} 

Mathematical reasoning has become a critical benchmark for evaluating and improving the capabilities of large language models (LLMs). Advances in this field have been driven by multiple factors, including architectural improvements \cite{mcleish2024transformersarithmeticrightembeddings}, enhanced pretraining datasets \cite{deepseekmathpushinglimitsmathematical,finemath,mathpile}, supervised fine-tuning \cite{yu2024metamathbootstrapmathematicalquestions,cobbe2021trainingverifierssolvemath}, reinforcement learning \cite{wang2024mathshepherdverifyreinforcellms,zelikman2022starbootstrappingreasoningreasoning}, and prompt-based methods such as chain-of-thought reasoning \cite{ye2024physics,auto-cot}. 
\citet{frieder2024largelanguagemodelsmathematicians} explored LLMs for assisting mathematicians, advocating a hybrid human-model approach. \citet{chang2023surveyevaluationlargelanguage,fan2024giantsshoulderseffortlessweak} evaluated LLMs in mathematical reasoning, noting strengths and limitations. \citet{Testolin_2024} and \citet{lu2023surveydeeplearningmathematical} analyzed deep learning in math problem-solving, highlighting challenges in generalization.

Despite advancements, LLMs in mathematical reasoning remain limited by reliance on dataset-driven learning, leading to brittleness and poor generalization \cite{ahn-etal-2024-large,lu2024twin,tian2025extrapolatingdecouplingimagetovideogeneration}. To address this, reinforcement learning has been employed to enhance verification mechanisms \cite{wang2024mathshepherdverifyreinforcellms}, while prompt engineering, such as physics-inspired prompting \cite{ye2024physics} and automated chain-of-thought generation \cite{auto-cot,fan2024giantsshoulderseffortlessweak}, has improved reasoning consistency. These findings highlight the need for structured reasoning techniques alongside architectural and data improvements to further advance mathematical capabilities in LLMs.

\subsection{Mathematical Datasets} 

Mathematical datasets for LLMs can be broadly categorized into numerical reasoning and automated theorem proving (ATP).
\textit{For numerical reasoning}, PRM800K \cite{lightman2024lets}, GSM8K \cite{cobbe2021trainingverifierssolvemath}, and GSM\_PLUS \cite{li-etal-2024-gsm} focus on arithmetic problem-solving, requiring step-by-step derivations. FormulaReasoning \cite{li2024formulareasoningdatasetformulabasednumerical} assesses formula-based numerical reasoning, while GAOKAO \cite{zhang2024evaluatingperformancelargelanguage} benchmarks LLMs' ability to solve complex mathematical problems in Chinese university entrance exams.
\textit{For automated theorem proving}, MiniF2F \cite{zheng2022miniff} compiles problems from formal proof assistants, including Metamath \cite{yu2024metamath}, Isabelle \cite{frieder2024largelanguagemodelsmathematicians}, and Lean \cite{han2022proof}. ProofNet \cite{azerbayev2023proofnet} spans undergraduate-level mathematics, bridging LLMs with formal proof verification. Additionally, DRAW-1K \cite{upadhyay-chang-2017-annotating} aids in equation derivation, while \citet{ying2024leanworkbooklargescalelean,wu2024leangithubcompilinggithublean} introduced datasets for Lean, supporting machine-verifiable proof generation.

In contrast, our STORM-BORN dataset focuses on challenging mathematical derivations in natural language, demanding complex reasoning and creativity, and is more likely to contain dense, human-like thinking patterns, such as approximations and heuristic cues.

\begin{figure*}[th]
    \centering
    \includegraphics[width=1.01\textwidth]{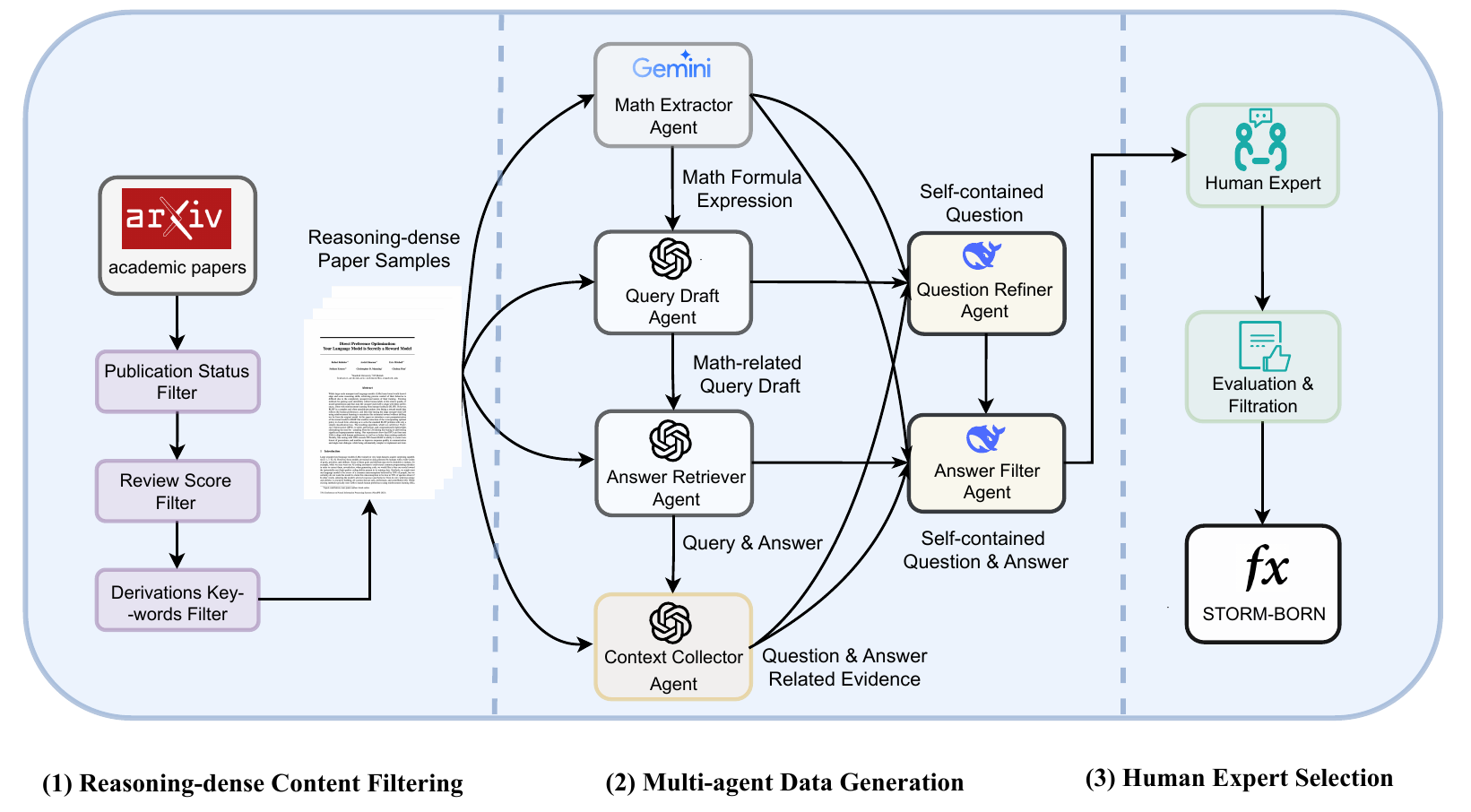}
\caption{Overview of the data generation framework of STORM-BORN, which consists of three main components: \textbf{(1) Reasoning-dense Content Filtering} selects reasoning-dense arXiv papers through linguistic markers and complexity criteria to ensure high-quality mathematical derivations. \textbf{(2) Multi-agent Data Generation} {orchestrates specialized agents for LaTeX extraction, query formulation, answer retrieval, and context enrichment, culminating in refined mathematical problems. } \textbf{(3) Human Expert Selection} applies rigorous evaluation criteria to select the most challenging and well-structured problems, resulting in the final STORM-BORN dataset for advancing mathematical reasoning capabilities.}
\label{fig:multi-agent-framework}
\end{figure*}

\section{Overall Pipeline}
In order to enhance LLMs’ reasoning abilities for mathematical expressions found in research papers, we created \textbf{STORM-BORN}, a dataset that involves advanced mathematical reasoning. This section describes in detail the construction process of \textbf{STORM-BORN}.

\subsection{Reasoning-dense Content Filtering}
\label{sec:content_filtering }
Distinguishing between basic concept explanations and genuinely complex reasoning requires human-like cognitive processes. To ensure our dataset contains more data and of higher quality, a key aspect lies in the selection of data sources—academic papers. Different papers vary in the amount and quality of data they provide, with some containing extensive mathematical content and detailed proofs and derivation processes, while others do not. Therefore, the focus should be on papers that not only contain a sufficient number of formulas but also provide thorough theorem proofs and derivation processes. More specifically, we select papers based on the following principles.

\paragraph{Publication Status and Review Score.} 
To ensure data reliability, we prioritized papers from reputable journals and conferences, which were peer-reviewed and met stringent acceptance criteria. We also limited the selection to papers published from May 2023 to October 2024 to ensure content freshness and reduce the risk of using outdated material. Additionally, all selected papers had to receive a score higher than "weak accept" from reviewers on the OpenReview platform, ensuring high data quality.

\paragraph{Richness of Mathematical Derivations.} 
We use linguistic markers such as ``assume'', ``derive'', and ``proof'' to filter papers that contain detailed derivations and complete sequences of proofs (most of this data comes from the appendices). If the target keywords appear more than five times in a paper, we consider it to have a higher likelihood of being our target paper. This ensures that the filtered papers contain high-quality mathematical reasoning.

\subsection{Multi-agent Data Generation}
\label{sec:data_processing}
We present a six-agent methodology to generate data. This streamlined workflow (see \cref{fig:multi-agent-framework}) ensures that each mathematical expression is accompanied by a coherent proof or derivation, a self-contained question, and a human-like step-by-step answer (a detailed explanation of this process is included below). We spent 200 USD on GPT-o1-Pro and spent about three weeks on prompt engineering. \cref{sec:workload} contains further details.
This multi-agent framework aims to generate high-quality mathematical data by systematically extracting expressions, posing meaningful questions, retrieving and refining answers, gathering requisite background information, and presenting the self-contained results, ultimately providing more transparent insight into mathematical derivations and proofs. In each step, all mathematical symbols and expressions are converted to LaTeX format.
\paragraph{Why not single-agent?}
We initially experimented with a single-agent approach for data generation, but the results were poor. The task is inherently complex and involves multiple steps. Using a single LLM leads to excessively long prompts with numerous critical points, making it difficult for the model to follow the instructions effectively. By employing a multi-agent system, we can decompose the task into smaller, more manageable components, allowing each LLM agent to focus on a specific step or key point, which improves the results. Additionally, this modular approach provides greater flexibility, making it easier to modify, refine, or integrate new modules for further improvements. In practice, the multi-agent system significantly enhances both the efficiency and quality of data generation.

\paragraph{Math Expression Extractor Agent}
We utilize lightweight multi-modal LLMs with extensive prompts for accurate LaTeX formula extraction, avoiding the limitations of traditional OCR techniques \cite{he-etal-2024-olympiadbench}. It uses a multi-modal large language model (MLLM) that can recognize mathematical expressions in text. After collecting these expressions, the original paper and the extracted expressions are forwarded to the Query Draft Agent.
    
\paragraph{Query Draft Agent} We employ the GPT-o1-Pro LLM as our Query Draft Agent, leveraging a well-structured and effective long prompt \textbf{exceeding 1k tokens}. It receives the entire paper rather than the chunked paper, which ensures it can comprehensively understand the entire paper. For each expression extracted from the Math Expression Extractor Agent, it generates at least one \emph{query}, focusing on the theorem or formula derivation problems. We also add a few shots to enhance the output format stability. The details of its prompt are in \cref{prompt:Query Draft}.

\paragraph{Answer Retriever Agent} 
The Answer Retriever takes the entire paper, a given expression, and its corresponding \emph{query} as input. The Answer Retriever Agent searches the paper for relevant content that can answer the query. Once relevant content is found, it extracts the entire answer directly from the paper rather than make a proof itself to avoid hallucination. Similar to Query Draft Agent, practice has proved that the task of this agent is also difficult and requires a more powerful LLM (\eg GPT-o1-Pro). The effective prompt we finally use is also relatively long with \textit{nearly 500 tokens}. The details of this prompt are in \cref{prompt:Answer Retriever}.

\paragraph{Context Collector Agent} 
Although Query Draft Agent and Answer Retriever Agent could generate high-quality \emph{query} and \emph{answer}, there still remains the possibility that they lack full information to make them self-contained, which means that the LLMs and humans could answer the question without reading the original paper. The Context Collector captures this information and stores it as \emph{evidence} for the target self-contained \emph{questions} and \emph{answers}.

\paragraph{Question Refiner Agent} 
 The goal of this agent is to incorporate the information from the \emph{evidence} into the \emph{query} and \emph{answer}, thereby generating \emph{self-contained question} that can be answered independently without reading the original resource.

\paragraph{Answer Filter Agent} 
Since our goal is to focus on mathematical reasoning, the Answer Filter Agent filters out any irrelevant content after receiving the data processed by the Question Refiner Agent, retaining only the essential information needed to understand how the expression is derived or proven. By filtering out unnecessary data, the subsequent modules can significantly reduce redundant workload and generate self-contained questions and answers.

\subsection{Human Expert Selection}

\label{sec:human_select}
Through Multi-agent Data Generation, we obtained 2k samples. We could have directly retained these 2k samples, but our goal was to extract the most challenging and high-quality dataset. To achieve this, we employed a group of expert mathematicians to conduct a rigorous selection process, ultimately arriving at a refined set of 100 samples. 
We sent the self-contained question and answer generated in (Sec.~\ref{sec:data_processing}) to human experts who are familiar with the reasoning-dense paper samples for selection. Human experts conducted strict audits on data quality, retained data that meets the standards, eliminated data that has no research value, and manually modified and optimized data that was not of borderline quality but could be improved.
Each paper was processed by experts for about 30 samples of questions and answers, and the processing of a single paper took about 15 minutes.
Through iterative expert feedback and revision, we refined the dataset, ensuring that each sample met the high-quality standards set by our guiding principles. 
This expert-driven process was critical to ensuring that the dataset reflects complex human-like mathematical reasoning, resulting in the final \textbf{STORM-BORN} dataset. This process was guided by the following five core principles: Reasoning Type, Problem Clarity, Derivation Correctness and Reasoning Density.

\paragraph{(Q1) Reasoning Type:} \textit{Does the problem demand creative insight and complex reasoning?} Initially, mathematicians determine whether the problem involves genuinely complex reasoning like deriving or proving a formula, as opposed to simple explanation or definition. 

\paragraph{(Q2) Problem Clarity:} \textit{Is the problem clear, well-defined, and solvable with the existing information?} This step evaluates the explicitness of the problem’s goal and conditions. Ambiguities or incomplete queries, where critical context is missing, are flagged for refinement. Human expert intervention is crucial here, as mathematical clarity often requires subjective interpretation, especially when key information is implied or subtly conveyed.

\paragraph{(Q3) Derivation Correctness:} \textit{Are all derivation steps logically valid, error-free, and complete?} Mathematicians carefully review each derivation step for correctness, ensuring that all logical transitions are accurate and coherent. This stage presents a significant challenge, as identifying logical errors or omissions often requires a deep theoretical understanding and specialized expertise.

\paragraph{(Q4) Reasoning Density:} \textit{Does the reasoning process include sufficient logical steps, exhibit heuristic reasoning cues, and demonstrate trial-and-error similar to human problem-solving?} This requires human expertise to assess whether the reasoning is sufficiently dense, complete, and heuristic. Mathematicians identify patterns in the reasoning that reflect human-like trial-and-error approaches. Missing or incomplete justifications are flagged for further revision.

\section{Experiments}\label{sec:experiments}

\begin{figure*}[tbp]
\centering
\includegraphics[width=\textwidth]{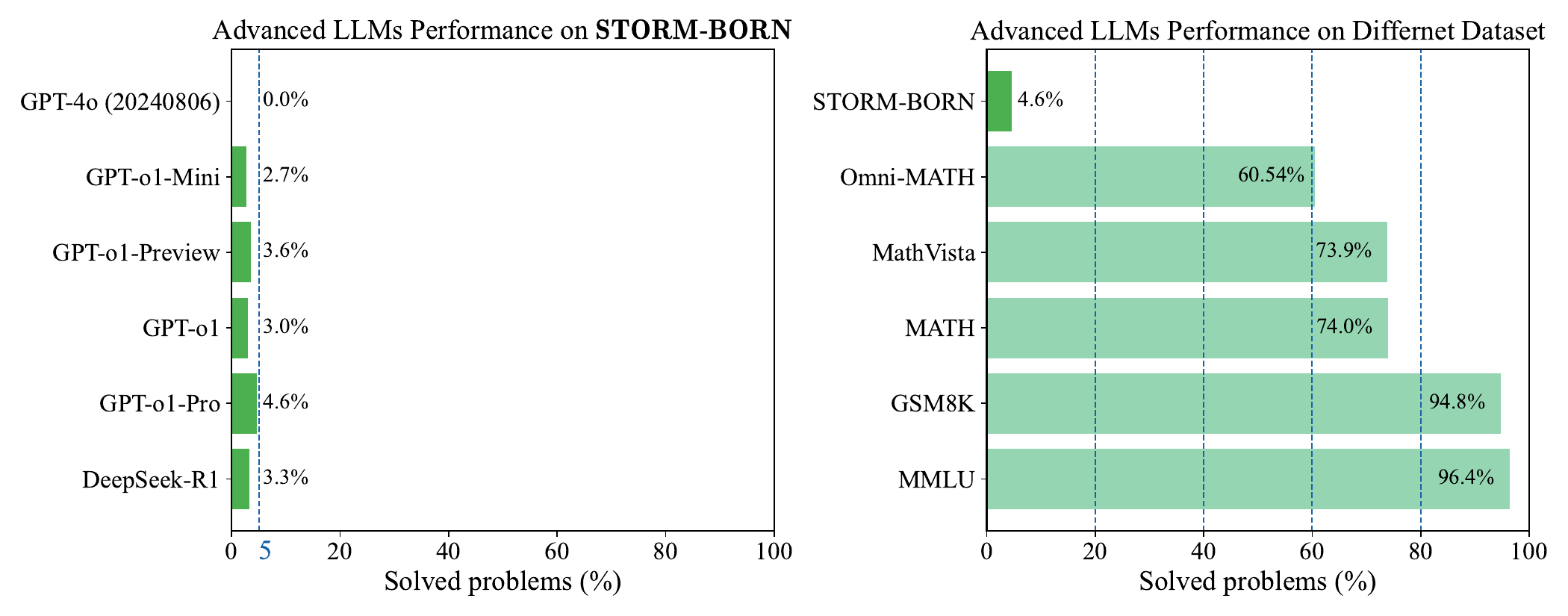}\\
\caption{{Performance of leading language models on STORM-BORN based on a human expert evaluation. All models show consistently poor performance, with even the best models solving less than 5\% of problems. When re-evaluating problems that were solved at least once by any model, GPT-o1-Pro demonstrated the strongest performance across
repeated trials.}}
\label{fig:STORM-BORN-solved-problem}
\end{figure*}

\definecolor{mygreen}{RGB}{0, 147, 123} 
\definecolor{mygrey}{RGB}{128,128, 128}
\begin{table*}[th]
\centering
\resizebox{0.9\linewidth}{!}{%
\begin{tabular}{lllllll}
\toprule
\textbf{Model} & \textbf{GSM8K-0shot} & \textbf{GSM8K-8shot}  & \textbf{MATH-0shot} & \textbf{MATH-4shot} \\
\midrule
\textbf{Tiny-LLaMA-1.1B-Chat} &  ~~1.36\% &  ~~2.19\% &  ~~1.20\% &     ~~2.14\% \\
\textbf{Tiny-LLaMA-1.1B-Chat} {\footnotesize (Ours)} & \textbf{~~2.05\%} {\color{mygreen}\footnotesize ($\uparrow$ 0.69)} & \textbf{~~2.65\%} {\color{mygreen}\footnotesize ($\uparrow$ 0.46)} & \textbf{~~4.00\%} {\color{mygreen}\footnotesize ($\uparrow$ 2.80)} & ~~\textbf{3.46}\% {\color{mygreen}\footnotesize ($\uparrow$ 1.32)}\\
\textbf{LLaMA2-7B} & ~~7.96\% & 14.33\% & ~~1.60\% & ~~4.44\%\\
\textbf{LLaMA2-7B} {\footnotesize (Ours)} & \textbf{~~8.80\%} {\color{mygreen}\footnotesize ($\uparrow$ 0.84)} & \textbf{16.98\%} {\color{mygreen}\footnotesize ($\uparrow$ 2.05)} & \textbf{~~2.60\%} {\color{mygreen}\footnotesize ($\uparrow$ 1.00)} & \textbf{~~4.56\%} {\color{mygreen}\footnotesize ($\uparrow$ 0.12)}\\
\textbf{LLaMA3-8B} & 15.39\% & 50.27\% &~~0.12\%  & 17.08\%\\
\textbf{LLaMA3-8B} {\footnotesize (Ours)} & \textbf{42.91\%} {\color{mygreen}\footnotesize ($\uparrow$ 27.52)} & {45.49\%} {\color{mygrey}\footnotesize ($\downarrow$ 4.78)} & \textbf{~~7.96\%} {\color{mygreen}\footnotesize ($\uparrow$ 7.84)} & {13.82\%} {\color{mygrey}\footnotesize ($\downarrow$ 3.26)}\\
\textbf{Qwen2.5-7B} &80.67\% & 83.32\%& 67.82\% & 54.42\% \\
\textbf{Qwen2.5-7B} {\footnotesize (Ours)} & \textbf{81.96\%} {\color{mygreen}\footnotesize ($\uparrow$ 1.29)} & \textbf{83.70\%} {\color{mygreen}\footnotesize ($\uparrow$ 0.38)} & \textbf{67.96\%} {\color{mygreen}\footnotesize ($\uparrow$ 0.14)}& \textbf{63.54\%} {\color{mygreen}\footnotesize ($\uparrow$ 9.12)} \\
\bottomrule
\end{tabular}
}
\caption
{Experimental results of four LLMs on GSM8K and MATH. Models marked with ``(Ours)'' are additionally fine-tuned on the STORM-BORN dataset.  The uparrow ($\uparrow$) beside each score denotes the absolute accuracy gain obtained after this fine-tuning, relative to the corresponding base model.
}
\label{tab:performance}
\vspace{-0.4cm}
\end{table*}

\begin{table}[t]
    \centering
    \resizebox{0.9\columnwidth}{!}{%
    \begin{tabular}{lllllll}
    \toprule
    \textbf{Model} & \textbf{AIME2024} & \textbf{AIME2025} \\
    \midrule
    \textbf{Qwen2.5-7B} & ~~0.00\% & ~~3.33\% \\
    \textbf{Qwen2.5-7B} {\footnotesize (Ours)} & \textbf{ ~3.33\%} {\color{mygreen}\footnotesize ($\uparrow$ 3.33)} & \textbf{ ~6.67\%} {\color{mygreen}\footnotesize ($\uparrow$ 3.34)}  \\
    \textbf{Qwen2.5-32B}  & {20.00\%}  & {15.00\%}  \\
    \textbf{Qwen2.5-32B} {\footnotesize (Ours)} & \textbf{23.33\%} {\color{mygreen}\footnotesize ($\uparrow$ 3.33)} & {15.00\%} {\color{mygreen}\footnotesize ($\uparrow$ 0.00)} \\
    \bottomrule
    \end{tabular}
    }
    \caption{
    Accuracy (\%) of Qwen 2.5 7B and 32B on the AIME 2024 and AIME 2025. Models marked with ``(Ours)'' are additionally fine-tuned on the STORM-BORN dataset. The uparrow ($\uparrow$) beside each score denotes the absolute accuracy gain obtained after this fine-tuning, relative to the corresponding base model.
    }
    \label{tab:aime}
\vspace{-0.4cm}
\end{table}

\begin{table}[t]
\centering
\label{tab:method_comparison}
\resizebox{1.0\linewidth}{!}{%
\begin{tabular}{lcccc}
\toprule
\textbf{Method} & \textbf{Correctness} & \textbf{Completeness} & \textbf{Similarity} & \textbf{Avg.} \\
\midrule
\textbf{Qwen2.5-7B} & 1.10 & 1.14 & 0.50 & 0.91 \\
\textbf{Qwen2.5-7B} (Ours)& \textbf{1.23} & \textbf{1.32} & \textbf{0.66} & \textbf{1.07} \\
\bottomrule
\end{tabular}
}
\caption{Formula Derivation Performance}
\end{table}

\subsection{Case Study}
\label{subsec:case_study}
In this preliminary case study, we compared three different types of datasets (see \cref{fig:splash}):
(I) Numerical reasoning datasets such as PRM-800K, which mainly examine numerical calculations, but may be too simple for advanced language models. For example, it can be solved like the expected value of a coin toss, which first calculates the probability of heads and tails, then calculates the payoff.  (II) Formal proof datasets, such as Minif2F, which use formal languages such as Lean to describe problems. Although rigorous, they are not easy to understand intuitively and are not easy to associate with real-world scenarios. Moreover, the answer examples can be solved with only one ring. (III) Our proposed STORM-BORN dataset focuses more on human-like reasoning processes and requires deeper understanding, flexible thinking, and complex reasoning. It is not only more challenging than (I), but also more interpretable and general than (II).  We selected the DPO \cite{rafailov2023direct} paper as our example, and then the system captured the derivation of important formulas and extracted the complete details of the derivation from the appendix of the paper, which added some human expert thinking, demonstrating the effectiveness of our method in scenarios of complex research.

\subsection{Human Evaluation}
\label{subsec:human_eval}

To validate the challenges of our dataset, we use STORM-BORN as a benchmark and evaluate current state-of-the-art models. Given that our data includes complex mathematical proofs, direct correctness evaluation is difficult. We observe that LLMs often overrate derivation accuracy (\eg DeepSeek-R1 assigns a perfect score to flawed cases). This makes expert validation essential since mathematical derivations differ fundamentally from numerical reasoning or formal proofs.

In our experiments, we identified consistent error patterns in LLMs, such as algebraic miscalculations, unproven assumptions, symbol misuse, and unjustified logical leaps. Detecting these problems requires reliable expert annotation. To ensure robustness, we conducted multiple evaluation runs, with each model generating three responses per question, independently scored by experts. A fully correct derivation receives 1 point, while partial credit reflects the proportion of key steps completed. We evaluated six leading language models: GPT-o1-Pro, GPT-o1, GPT-o1-Preview, GPT-4o, and DeepSeek-R1. As shown in \cref{fig:STORM-BORN-solved-problem}, the most advanced models (GPT-o1-Pro) achieve just 5\% accuracy on STORM-BORN versus 96.4\% on MMLU, demonstrating our dataset's unique challenge to mathematical reasoning.

\subsection{Automatic Evaluation}

To evaluate the fine-tuning effectiveness of our dataset, we assess it in two key aspects: in-domain formula derivation performance and out-of-domain numerical reasoning generalization ability.

\begin{table*}[t]
\centering
\resizebox{0.75\linewidth}{!}{%
\begin{tabular}{lcccc}
\toprule
\textbf{Method} & \textbf{GSM8k-0shot} & \textbf{GSM8k-8shot} & \textbf{Math-0shot} & \textbf{Math-4shot} \\
\midrule
\textbf{LLaMA2-7B} & 7.96 & 14.33 & 1.60 & 4.44 \\
\textbf{LLaMA2-7B + top-100} & \textbf{8.80} & \textbf{16.98} & \textbf{2.60} & \textbf{4.56} \\
\textbf{LLaMA2-7B + top-500} & 8.11 & 15.09 & 2.40 & 4.52 \\
\textbf{LLaMA2-7B + 2k} & 6.70 & 14.87 & 2.28 & 4.46 \\
\textbf{LLaMA2-7B + fullset} & 5.16 & 14.10 & 2.58 & 4.92 \\
\bottomrule
\end{tabular}
}
\caption{Quality Ablation Performance}
\label{tab:performance_comparison}
\end{table*}

\paragraph{Numerical Reasoning}

To assess the generalization of STORM-BORN on mathematical reasoning, we conduct full fine-tuning experiments on four most popular models: TinyLLaMA-1.1B-chat \cite{zhang2024tinyllamaopensourcesmalllanguage}, LLaMA2-7B \cite{touvron2023llama2openfoundation}, LLaMA3-8B \cite{grattafiori2024llama3herdmodels} and Qwen2.5-7B \cite{qwen2025qwen25technicalreport}, evaluating their performance on GSM8K \cite{cobbe2021trainingverifierssolvemath}, MATH \cite{math} and AIME. For few-shot evaluation, we follow the setup from \citet{touvron2023llama2openfoundation}, using 8-shot for GSM8K and 4-shot for MATH. Due to the extreme difficulty of AIME and the negligible performance of LLaMA-based models (scoring near zero), we evaluate only Qwen2.5-7B on this benchmark.

As shown in Table~\ref{tab:performance}, fine-tuning solely on 100 STORM-BORN samples significantly boosts performance across benchmarks:relative improvement of 233.3\% MATH for Tiny-LLaMA-1.1B-chat, 62.5\% for LLaMA2-7B, absolute improvement of 7.84\% MATH for LLaMA3-8B, and 9.12\% for Qwen2.5-7B. These results demonstrate that STORM-BORN generalizes well to out-of-domain numerical reasoning tasks. Moreover, the improvement increases with model capability, indicating that stronger foundation models benefit more from our dataset. Table~\ref{tab:aime} illustrates that even on the highly challenging AIME benchmark, our dataset continues to drive improvements in model performance. This highlights the strong generalization and reasoning capabilities gained through fine-tuning on STORM-BORN, particularly for advanced problem-solving skills required in expert-level mathematics.

\paragraph{Formula Derivation}

We evaluated our method on a formula derivation test set extracted from NuminaMath-1.5 \cite{numina_math_datasets}. We fully fine-tune Qwen2.5-7B \cite{qwen2025qwen25technicalreport} on STORM-BORN and use Deepseek-R1 for evaluation. We assessed correctness, completeness, and similarity to ground-truth proofs (each scored on a 0–2 scale) by comparing both ground-truth proofs and predicted derivations. The evaluation prompt is detailed in Appendix~\ref{app:eval}. The results show a relative improvement of 17.58\%, with the average score increasing from 0.91 to 1.07. This demonstrates that our dataset directly benefits formula derivation reasoning.

\subsection{Quality Ablation}

Our dataset prioritizes quality and difficulty, which led us to reduce the initial 2,000 automatically generated samples down to a rigorously curated set of 100 high-quality examples. To evaluate how quality and quantity affect fine-tuning performance, we slightly relaxed our selection criteria (reasoning density), to construct larger subsets: top-100, top-500, and a full 2k automated set. We then conducted a comparative experiment across these subsets.

As shown in Table~\ref{tab:performance_comparison}, we can observe: (1) LLaMA2-7B + top-100 outperforms larger datasets: On GSM8K, it achieves higher accuracy in both zero-shot (8.80 vs. 8.11) and 8-shot (16.98 vs. 15.09) settings compared to the top-500 and 2k sets, demonstrating that small, high-quality data enhances reasoning more effectively than larger, lower-quality data. (2) Full unfiltered set harms performance: LLaMA2-7B trained on the full, uncurated set underperforms the base model (\eg GSM8K zero-shot: 5.16 vs. 7.96). In contrast, the 2k set generated by our pipeline without human validation still improves over the base model (\eg 14.87 vs. 14.33), confirming the effectiveness of our synthesis framework.

\section{Conclusion}

In conclusion, we present \textbf{STORM-BORN}, a novel dataset designed to address the limitations of existing mathematical derivation datasets. Curated from recent top-tier academic papers via the arXiv repository, STORM-BORN is both nuanced and scalable, while avoiding data contamination. Unlike isolated steps, we capture full derivations to preserve logical flow and encourage deep theoretical reasoning. Using a human-in-loop and multi-agent LLM framework \textit{STORM}, we generate problems requiring at least three reasoning steps, ensuring complexity and creativity. Expert evaluations ensure reliable annotations. Empirical results highlight the dataset's challenge, with advanced LLMs like GPT-o1-Pro solving fewer than 5\% of the problems, compared to 95\% accuracy on GSM8K. Additionally, STORM-BORN demonstrates strong generalization capabilities, offering a high-difficulty evaluation benchmark for AI's approach to mathematician-level reasoning.
In this way, STORM-BORN will be a pioneer in deriving expert evaluation and provide a human-evaluation and auto-evaluation template for future work.

\section*{Limitations}
This study addresses an important gap in the field, but it also faces certain limitations. Specifically, the automated evaluation of data quality remains challenging, as our focus on complex mathematical derivations rather than numerical computing makes quality assessment difficult (a problem also noted by \citet{glazer2024frontiermathbenchmarkevaluatingadvanced}). Currently, we rely primarily on a carefully designed multi-agent curation pipeline and manual inspection by mathematicians. However, with the rapid advancement and scaling of LLMs, we believe that in the future, LLMs can be fully employed to automate this process, iteratively improving and optimizing it.

\section*{Ethics Statement}
The dataset construction process in this study strictly adheres to ethical guidelines and fully complies with relevant legal regulations. We obtain publicly accessible, high-quality academic papers and utilize a combination of multimodal models and human evaluation feedback for data processing and optimization, ensuring data quality and reliability before generating the final dataset. The entire data collection and processing workflow is transparent and traceable, with all papers sourced from legal and publicly available channels, guaranteeing compliance and traceability of data.
The dataset constructed in this study is intended solely for academic research and experimental purposes, with no involvement in commercial applications or risk of sensitive information leakage. 

\bibliography{main}

\appendix
\onecolumn

\clearpage
\newpage

\section{Workload and Prompts}
\label{sec:workload}
We invested a lot of work, energy, and time in this research. Our goal is to generate high-quality formula derivation and question-answering. At first glance, this seems to be a simple task, but in fact it involves extremely complex and extensive workload. Initially, we explored various technical solutions, such as optical character recognition (OCR), but when using OCR for formula recognition and extraction, we often encountered incomplete positioning (only part of the formula was framed out), resulting in inaccurate formula extraction. After repeated comparisons and experiments, we finally chose the method of multi-agent large language model (LLM) collaboration, which has consumed some time and energy.

The biggest challenge appeared in the prompt design and optimization stage. Practice has shown that LLM will encounter a series of problems, such as identifying key data in long texts, following instructions, and producing stable output. To solve these difficulties, we continuously refined the overall workflow and assigned complex tasks to multiple appropriate numbers of agents (see \cref{fig:multi-agent-framework}) for collaborative execution. At the same time, the prompts of each agent were modified, iterated, and verified for multiple rounds. This process is tedious and time-consuming, and consumes a lot of energy.

Regarding manual evaluation and feedback, each paper required individuals with relevant academic background to read, assess, and provide feedback on the generated data, which increases labor and time costs.

For resource costs and time costs, please see \cref{costs}.

Thanks to this painstaking and systematic workflow, we were finally able to obtain high-quality question-answering data. We will introduce our prompts below, hoping to provide further insight into the complexity of this study, the extensive workload involved, and our efforts to overcome a variety of challenges.

\subsection{Math Expression Extractor Agent}
\label{prompt:Math Expression Extractor}
We encountered many problems in the process, such as: the set of extracted mathematical expressions omitted important items, contained unnecessary items and repeated items; the output latex format did not meet the requirements. To solve these problems, we added new rules to the prompt and repeatedly verified the effect in practice, \textbf{and iterated continuously}. Through repeated iterations in practice, these problems were solved, which enables the MLLM to follow the instructions to extract all important mathematical expressions (formulas, theorems, lemmas, etc.), ignore unimportant mathematical expressions (such as intermediate expressions that appear in the derivation process, mathematical content inserted in the paragraph), and ensure that the output expression is in the correct format.
\begin{tcolorbox}[breakable, colback=TiffanyBlue!5!white, colframe=TiffanyBlue!75!black, title={Prompt of Math Expression Extractor}]
\begin{lstlisting}
"""Read the paper, then:

1. Formula Recognition:
- Identify all mathematical formulas, theorems, lemmas, and corollaries in the paper. Especially Numbered formulas.Retain the formula's number (if any).
- For formulas without explicit labels (i.e., those not labeled as "theorem, " "lemma, " or "corollary"), classify them as "formula."
- Required types of formulas to recognize:
    - Numbered formulas.
    - Formulas that appear on separate lines (for example, occupying a line or multiple lines by themselves in the paper).
- Ignore:
    - Formulas that appear in the middle of a paragraph without separate lines or numbers.
- Make sure there are no duplicates in the results (duplicates refer to formulas that are exactly the same after conversion to LaTeX. If the same formula appears in the paper under different numbers, treat them as the same formula).

2. LaTeX Conversion (Convert the formulas identified in step 1 into LaTeX format strings):
- Symbols: Convert mathematical symbols accurately.
- Subscripts and superscripts: Convert subscripts and superscripts correctly.
- Uppercase and lowercase: Preserve the original variable and constant casing.
- Formula structure: Keep the entire structure of the formula intact.
- Formula numbering: Retain the formula's number (if any).
- Italics: For italicized variables in the text, wrap them with \textit{} in LaTeX.
- Math environment: Use `$ ... $` for inline formulas and `$$ ... $$` for block (display) formulas.
- Additional conditions: Check whether the paper includes definitions or explanations immediately following the formula (for example, "where X is ...") and incorporate them if present.

3. JSONL Output:
- Output all converted LaTeX strings in multi-line JSONL format so they can be parsed line by line.
- Each line should be a JSON object whose key is the type of the formula ("formula", "lemma", "theorem", "corollary", etc.) and whose value is the LaTeX string obtained from step 2.
- Be sure to follow the requirements in step 2!

Ensure the formulas are exactly the same as in the original text!"""
\end{lstlisting}
\end{tcolorbox}

\subsection{Query Draft Agent}
\label{prompt:Query Draft}
The more difficult task also leads to more problems encountered in the process, such as the generated questions are too rigid, the questions lack prerequisites, and only the formula reference number is output without the original formula {which emphasizes the need of Context Collector Agent and Question Refiner Agent.
\begin{tcolorbox}[breakable, colback=TiffanyBlue!5!white, colframe=TiffanyBlue!75!black, title={Prompt of Query Draft}]
\begin{lstlisting}
"""I will provide you with a dataset extracted from this paper, in JSONL format. Each entry is a dictionary whose keys are "formula, " "lemma, " "theorem, " etc., representing the category of the mathematical expression, and whose values contain a mathematical expression in LaTeX format, extracted from the paper.

Carefully read and understand the paper's content, especially the parts related to each formula in the JSONL. For each formula, please complete the following steps:

---

Step 1:
Locate where the formula is first defined or fully derived in the paper, and use the relevant context to extract all the direct necessary conditions for deriving or proving that formula. These preconditions include, but are not limited to:

1. Which other formulas this formula is derived from or depends on. For each such formula, record its full content (in LaTeX format), its numbering (if any), and its name (if any).
2. Relevant problem settings.
3. The specific meaning of symbols or variables involved in the formula.

---

Step 2:
Based on the extracted preconditions, generate a complete question that clearly asks how to derive or prove the formula. The question should include:

1. The formula itself: Present the full content of this formula (in LaTeX format). Do not only reference its number.
2. The preconditions: Explicitly integrate the preconditions extracted from the paper into the question. List out the full contents of all the formulas it depends on and reference them by their respective numbers or names. Do not produce a question such as "What are the preconditions?"

The form of the question must meet the following requirements:

- If a formula is derived from one or more other formulas, you must explicitly list the full content (in LaTeX) of these preceding formulas and reference them by their numbers or names, and explain how the current formula is derived from them. For example, if the paper contains Formula 3 (content: X) and Formula 4 (content: Y), and Formula 4 is derived from Formula 3, then the generated question should be:

"Based on Formula 3: X, how can we derive Formula 4: Y?"

- If the formula is a theorem, lemma, or corollary, please generate a question asking how to prove it, for example:

"How can we prove Lemma 1: X is true?"

Note: The question must be structured and logical, clearly showing the derivation or proof process of the formula and explicitly reflecting the dependency between formulas while fully presenting all related formulas.

---

Step 3:
Match each formula with its corresponding question and output the result in multi-line JSONL format.

Each data entry should be a dictionary containing the following two key-value pairs:

1. Formula type:
- The key is "formula, " "lemma, " "theorem, " etc.
- The value is the LaTeX content of the formula.
2. Generated question:
- The key is "query."
- The value is the complete question generated according to Step 1 and Step 2.

---

Important Notes:
1. Format Requirements:
- Ensure the output is in JSONL format, with each line corresponding to one data entry.
2. Formula Accuracy:
- If the question contains mathematical expressions, convert them into LaTeX format. Make sure they align with the original mathematical meaning. Minor formatting differences can be ignored.
3. LaTeX Conversion (Converts the mathematical expressions contained in the problem to strings in LaTeX format):
- Symbols: Convert mathematical symbols accurately.
- Subscripts and superscripts: Convert subscripts and superscripts correctly.
- Uppercase and lowercase: Preserve the original variable and constant casing.
- Formula structure: Keep the entire structure of the formula intact.
- Formula numbering: Retain the formula's number (if any).
- Italics: For italicized variables in the text, wrap them with \textit{} in LaTeX.
- Math environment: Use `$ ... $` for inline formulas and `$$ ... $$` for block (display) formulas.
4. Completeness of Preconditions:
- The question content must include all direct necessary conditions. Particularly, indicate which other formulas the current formula is derived from or depends on, and clearly specify the entire content, numbering, or name of those referenced formulas. Do not produce questions such as "What are the preconditions?"

---

Examples:
Here are some example questions and their corresponding output formats for reference:

- Suppose the paper contains the following formula:
{"lemma": "Lemma 1.  The function $f (x)$ is continuous."}
The generated question might be:
{"query":"How can we prove Lemma 1: The function $f (x)$ is continuous. is true?"}

- Suppose the paper contains the following formula:
{"formula": "y = mx + b"}
and it is explained that this formula is derived from y = f (x) and f (x) = mx + b. Then the generated question might be:
{"query":"Based on the formulas: $y = f (x)$ and $f (x) = mx + b$, how can we derive the formula: $y = mx + b$?"}

- Suppose the paper contains the following formula:
{"formula": "$$\\pi_r (y | x) = \\frac{1}{Z (x)} \\pi_{ref}(y | x) \\exp (\\frac{1}{\\beta} r (x, y))$$"}
and it is explained that this formula is derived from Formula 3, $KL (\\pi_r (y|x) || \\pi_{ref}(y|x)) \\leq \\epsilon$. Then the generated question should be:
{"query":"Based on Formula 3: $KL (\\pi_r (y|x) || \\pi_{ref}(y|x)) \\leq \\epsilon$, how can we derive Formula: $\\pi_r (y | x) = \\frac{1}{Z (x)} \\pi_{ref}(y | x) \\exp (\\frac{1}{\\beta} r (x, y))$?"}

The dataset is as follows:\n
\end{lstlisting}
\end{tcolorbox}

\subsection{Answer Retriever Agent}
\label{prompt:Answer Retriever}
In order to solve the problems encountered in the process, such as: \textbf{the answer is not extracted from the original text but the large model generates the answer itself}, the answer retrieved in this agent may lack the important complete proof process in the appendix, or is a summary of the answer in the original text, the effective prompt we finally get is also relatively long with nearly 500 tokens.
\begin{tcolorbox}[breakable, colback=TiffanyBlue!5!white, colframe=TiffanyBlue!75!black, width=\textwidth, title={Prompt of Answer Retriever}]
\begin{lstlisting}
"""I will provide a JSONL-format dataset extracted from this paper. Each piece of data in the dataset is a dictionary containing two main key-value pairs:
1. **Formula-related keys ("formula", "lemma", "theorem", etc.)** indicating the type of mathematical expression; the value is the LaTeX-formatted mathematical expression extracted from the paper.
2. **query**, whose value is a question generated by a large model based on the paper and the mathematical expression.

Please process this dataset according to the following steps and requirements.

---

### Step One:
For the "expression" and "query" in each piece of data, determine whether the answer to that question can be found in the paper. The specific steps are as follows:

1. **Find the first occurrence**
   - Locate where the expression first appears in the paper and check the surrounding context for relevant clues.
   - If there are any references or citations, follow those as well.

2. **Check the appendix and other sections**
   - Search the paper's appendix or other relevant chapters to see if the proof or derivation steps for that expression are provided. This may well be the answer to the question.

3. **Confirm feasibility**
   - If the paper does not include any relevant content addressing the question, you may skip this expression and proceed to the next one.
   - If the paper does indeed contain content that can answer the question, extract the relevant content from the original text.

When extracting the answer, please note the following requirements:
- **Completeness**: The extracted answers should cover all the relevant steps needed to solve the problem in the paper.
- **Consistency**: Include only content from the original text in the answer (you may make minimal necessary edits for coherence, but do not change the original meaning). Avoid adding extra content or descriptions not found in the original text.
- **Citation handling**: If the answer cites other formulas or theorems from the paper, also include their original content in the derivation or proof process, rather than leaving only references or labels.
- **LaTeX conversion**: Ensure all mathematical expressions are converted to the same LaTeX format as in the original text, including:
  - Accuracy of symbols, subscripts, superscripts, and capitalization.
  - Preserving the original structure and numbering (if any).
  - Using \textit{} for italicized variables.
  - Using $...$ for inline math expressions and $$...$$ for display math expressions.

---

### Step Two:
Match the answers extracted in Step One with the corresponding entries in the dataset, and add a new key-value pair to form a new data record. The specific requirements are:

- For each original data entry, add a new key called `whole_label`, whose value is the LaTeX-formatted answer content extracted from the paper.
- Output format must be **multi-line JSONL**, one piece of data per line:
  1. The original two key-value pairs remain unchanged and must not be modified.
  2. Add the `whole_label` key as the third key-value pair.

---

### Output Requirements:
1. **Multi-line JSONL format**: One data entry per line.
2. **Accuracy of content**: Formulas must match the original text of the paper exactly, with correct symbols, subscripts, superscripts, and capitalization.
3. ** Content consistency ** : Only retain the original content in the answer (you can make a small amount of necessary cohesive editing, but do not change the original meaning), and try to avoid adding additional content or descriptions that do not appear in the original.
---

### Note:
- Please strictly follow the above requirements to avoid omitting any key content.
- Ensure there are no errors or incomplete parts in the output text.

---

Below is the dataset:
""" 
\end{lstlisting}
\end{tcolorbox}

\section{Resource and Time Costs}\label{costs}
At the outset, it is important to highlight the considerable workload entailed in our approach, with the associated resource and time costs reflecting the extensive efforts required for its implementation.
\subsection{Resource Costs}
\label{costs:resource}
To support the multi-agent system, we subscribed to GPT-o1-Pro for a one-month period at an approximate cost of 200 USD. For simpler tasks, such as Math Expression Extraction and Answer Filtering, we utilize free LLMs as agents. For more complex tasks, such as Query Generation and Answer Retrieval, we rely on the paid GPT-o1-Pro model to ensure enhanced performance and accuracy.
An illustrative example is provided below. After the Math Expression Extraction step, using GPT-o1-Pro for both Query Generation and Answer Retrieval yielded the results shown in \cref{fig:o1-pro outputs}. In contrast, using the gemini-2.0-flash-exp model for the same tasks resulted in \cref{fig:gemini outputs1}. Furthermore, when using the query generated by GPT-o1-Pro and applying gemini-2.0-flash-exp for Answer Retrieval, the result was \cref{fig:gemini outputs2}. This comparison clearly illustrates the necessity of using more powerful models like GPT-o1-Pro for complex tasks.
\subsubsection{GPT-o1-Pro Results}
\label{fig:o1-pro outputs}
Here we provide the results of GPT-o1-Pro as follows:
\begin{tcolorbox}[breakable, colback=TiffanyBlue!5!white, colframe=TiffanyBlue!75!black, title={GPT-o1-Pro as Query Draft and Answer Retriever.}]
\textbf{formula:} \\[-0.35cm]
\begin{lstlisting}
"$$\\pi_{r}(y | x)=\\frac{1}{Z (x)} \\pi_{r e f}(y | x) \\exp (\\frac{1}{\\beta} r (x, y)), $$(4)"
\end{lstlisting}
\tcbline
\textbf{query:} \\[-0.35cm]
\begin{lstlisting}
"Based on Formula (3): $$\\max _{\\pi_{\\theta}} \\mathbb{E}_{x \\sim D, y \\sim \\pi_{\\theta}(y|x)}[r_{\\phi}(x, y)]-\\beta D_{K L}[\\pi_{\\theta}(y | x) || \\pi_{r e f}(y | x)], $$ which enforces a KL-constrained reward maximization, how can we derive Formula (4): $$\\pi_{r}(y | x)=\\frac{1}{Z (x)} \\pi_{r e f}(y | x) \\exp\\Bigl (\\tfrac{1}{\\beta} r (x, y)\\Bigr), $$ where $Z (x)$ is the partition function ensuring $\\pi_{r}$ is a valid probability distribution?"
\end{lstlisting}
\tcbline
\textbf{answer:} \\[-0.35cm]
\begin{lstlisting}
"In the paper's Appendix A.1 (pp. 15-16), ``Deriving the Optimum of the KL-Constrained Reward Maximization Objective, '' the authors write:\n\n``[We] optimize the following objective:\n$$\\max_{\\pi}\\, \\mathbb{E}_{x\\sim D, y\\sim\\pi (y\\mid x)}\\bigl[r (x, y)\\bigr]\\; -\\; \\beta\\, D_{\\mathrm{KL}}\\bigl[\\pi (y\\mid x)\\, \\|\\, \\pi_{\\mathrm{ref}}(y\\mid x)\\bigr].$$\n\n... We now have:\n$$\\max_{\\pi}\\, \\mathbb{E}_{x\\sim D, y\\sim \\pi (y\\mid x)}\\bigl[r (x, y)\\; -\\; \\log\\frac{\\pi (y\\mid x)}{\\pi_{\\mathrm{ref}}(y\\mid x)}\\bigr]\\; =\\; \\max_{\\pi}\\, \\mathbb{E}_{x\\sim D, y\\sim \\pi (y\\mid x)} \\bigl[\\, r (x, y)\\; -\\; \\log\\pi (y\\mid x)\\; +\\; \\log\\pi_{\\mathrm{ref}}(y\\mid x)\\bigr].$$\n\nSince $Z (x)$ does not depend on $\\pi$, the minimum is achieved by the policy that minimizes the first KL term. Gibbs' inequality tells us that the KL-divergence is minimized at 0 if and only if the two distributions are identical. Hence we have the optimal solution:\n$$\\pi^{*}(y\\mid x)\\; =\\; \\frac{1}{Z (x)}\\, \\pi_{\\mathrm{ref}}(y\\mid x)\\; \\exp\\Bigl (\\tfrac{1}{\\beta}\\, r (x, y)\\Bigr), $$\nfor all $x\\in\\mathcal{D}$. This completes the derivation.''\n\n\\(\\textit{Here, }\\; Z (x)=\\sum_{y}\\, \\pi_{\\mathrm{ref}}(y\\mid x)\\exp\\bigl (\\tfrac{1}{\\beta}r (x, y)\\bigr)\\; \\textit{ensures}\\; \\pi^{*}(y\\mid x)\\; \\textit{is a valid probability distribution.}\\)\n\nevidence: Appendix A.1, pp. 15-16 of the paper"}
\end{lstlisting}
\end{tcolorbox}

\subsubsection{Gemini-2.0-Flash-Exp Results}
\label{fig:gemini outputs1}
Here we provide the results of Gemini-2.0-Flash-Exp as follows:
\begin{tcolorbox}[breakable, colback=TiffanyBlue!5!white, colframe=TiffanyBlue!75!black, title={Gemini-2.0-Flash-Exp as Query Draft and Answer Retriever, using the same formula as \cref{fig:o1-pro outputs}.  It is evident that compared to \cref{fig:o1-pro outputs}, the query generated by gemini-2.0-flash-exp did not fully display the formula conditions (formula 4 in \cref{fig:splash}), and the derivation process was overly general rather than a detailed step-by-step derivation.}]
\textbf{formula:} \\[-0.35cm]
\begin{lstlisting}
"$$\\pi_{r}(y | x)=\\frac{1}{Z (x)} \\pi_{r e f}(y | x) \\exp (\\frac{1}{\\beta} r (x, y)), $$(4)"
\end{lstlisting}
\tcbline
\textbf{query:} \\[-0.35cm]
\begin{lstlisting}
"Given the RL objective in Formula (3):$$\\max _{\\pi_{\\theta}} \\mathbb{E}_{x \\sim D, y \\sim \\pi_{\\theta}(y|x)}[r_{\\phi}(x, y)]-\\beta D_{K L}[\\pi_{\\theta}(y | x) || \\pi_{r e f}(y | x)], $$ how can we express the optimal solution $\\pi_{r}(y | x)$ to the KL-constrained reward maximization objective, where $Z (x)$ is the partition function?"
\end{lstlisting}
\tcbline
\textbf{answer:} \\[-0.35cm]
\begin{lstlisting}
"Following prior work [31, 30, 19, 15], it is straightforward to show that the optimal solution to the KL-constrained reward maximization objective in Eq. 3 takes the form:\n$$\\pi_{r}(y | x)=\\frac{1}{Z (x)} \\pi_{r e f}(y | x) \\exp (\\frac{1}{\\beta} r (x, y)), $$(4)\nwhere $$Z (x) = \\sum_{y} \\pi_{r e f}(y | x) \\exp (\\frac{1}{\\beta} r (x, y)), $$ is the partition function. See Appendix A.1 for a complete derivation."
\end{lstlisting}
\end{tcolorbox}

\subsubsection{Gemini-2.0-Flash-Exp Results}
\label{fig:gemini outputs2}
Here we provide the results of Gemini-2.0-Flash-Exp as follows:
\begin{tcolorbox}[breakable, colback=TiffanyBlue!5!white, colframe=TiffanyBlue!75!black, title={Gemini-2.0-Flash-Exp as Answer Retriever, using the same formula and query as \cref{fig:o1-pro outputs}. It is evident that the result was less comprehensive and detailed than \cref{fig:o1-pro outputs}.}]
\textbf{formula:} \\[-0.35cm]
\begin{lstlisting}
"$$\\pi_{r}(y | x)=\\frac{1}{Z (x)} \\pi_{r e f}(y | x) \\exp (\\frac{1}{\\beta} r (x, y)), $$(4)"
\end{lstlisting}
\tcbline
\textbf{query:} \\[-0.35cm]
\begin{lstlisting}
"Based on Formula (3): $$\\max _{\\pi_{\\theta}} \\mathbb{E}_{x \\sim D, y \\sim \\pi_{\\theta}(y|x)}[r_{\\phi}(x, y)]-\\beta D_{K L}[\\pi_{\\theta}(y | x) || \\pi_{r e f}(y | x)], $$ which enforces a KL-constrained reward maximization, how can we derive Formula (4): $$\\pi_{r}(y | x)=\\frac{1}{Z (x)} \\pi_{r e f}(y | x) \\exp\\Bigl (\\tfrac{1}{\\beta} r (x, y)\\Bigr), $$ where $Z (x)$ is the partition function ensuring $\\pi_{r}$ is a valid probability distribution?"
\end{lstlisting}
\tcbline
\textbf{answer:} \\[-0.35cm]
\begin{lstlisting}
"Following prior work [31, 30, 19, 15], it is straightforward to show that the optimal solution to the KL-constrained reward maximization objective in Eq. 3 takes the form:\n$$\\pi_{r}(y | x)=\\frac{1}{Z (x)} \\pi_{\\textit{ref}}(y | x) \\exp (\\frac{1}{\\beta} r (x, y)), $$(4)\nwhere $Z (x)=\\sum_{y} \\pi_{\\textit{ref}}(y | x) \\exp (\\frac{1}{\\beta} r (x, y))$ is the partition function. evidence: Section A.1, Appendix"
\end{lstlisting}
\end{tcolorbox}
\subsection{Time Costs}
On average, the processing time for each paper using the multi-agent system is approximately 20 minutes, with variability depending on the paper's length and the number of formulas. When utilizing GPT-o1-Pro for more challenging tasks, the processing time can be significantly longer. Moreover, output failures may occur, requiring multiple retries—sometimes two or even three times—leading to substantial time costs. Additionally, issues such as the ``dumbing down'' of LLMs during intensive tasks can further hinder experimental progress, creating delays in task completion. This represents a significant source of time cost, as repeated attempts are necessary to recover from failures and ensure valid results. In addition, there is no API for GPT-o1-Pro, so we have to use the web version. And the model can not receive pdf files, so we can only convert the paper into page screenshots and gradually upload, which increases the labor costs and time costs.

\section{Derivation Evaluation Prompt}\label{app:eval}

We evaluated our method on a formula derivation test set using Deepseek-R1 with the prompt below to assess correctness, completeness, and similarity to ground-truth proofs (each scored on a 0–2 scale). We found that providing the ground-truth proof alongside the model's output was crucial for accurate comparison and reliable scoring.

\begin{tcolorbox}[breakable, colback=TiffanyBlue!5!white, colframe=TiffanyBlue!75!black, width=\textwidth, title={Prompt of Answer Retriever}]
\begin{lstlisting}
"""
You are a precise mathematical proof evaluator for proof problems. The user will provide both a problem statement, ground truth proof and a proposed solution. 
    Evaluate the solution based on the ground truth proof, score the solution based on the following criteria:

    1. **Correctness (0-2):**
    - 0: Fundamentally wrong.
    - 1: Partially correct with significant flaws.
    - 2: Fully correct and logically sound.

    2. **Completeness (0-2):**
    - 0: Incomplete; key steps are missing.
    - 1: Moderately complete; some steps or justifications are missing.
    - 2: Fully complete; all necessary steps and justifications are present.

    3. **Similarity (0-2):**
    - 0: No similarity; completely different from the ground truth.
    - 1: Some similarity; some steps or justifications are similar.
    - 2: High similarity; all steps and justifications are identical.

    Output your evaluation as a JSON object in the format:
    {"correctness": <0-2>, "completeness": <0-2>, "similarity": <0-2>}
""" 
\end{lstlisting}
\end{tcolorbox}

\end{document}